\documentclass[runningheads]{llncs}
\usepackage[T1]{fontenc}
\usepackage{url}
\usepackage{graphicx}
\usepackage{amssymb}
\usepackage{amsmath}
\usepackage{subfigure}
\usepackage{booktabs}

\usepackage[pagebackref=true,breaklinks=true,letterpaper=true,colorlinks,citecolor=blue,linkcolor=blue,bookmarks=false]{hyperref}
\usepackage{multirow}
\begin{document}
\title{U-shaped Vision Mamba for Single Image Dehazing}
%
%
\author{Zhuoran Zheng\inst{1} \and
Chen Wu\inst{2} 
}
%
%
\institute{Nanjing University of Science and Technology \and
University of Science and Technology of China \\
\email{zhengzr@njust.edu.cn}}
\maketitle              
\begin{abstract}

Currently, Transformer is the most popular architecture for image dehazing, but due to its large computational complexity, its ability to handle long-range dependency is limited on resource-constrained devices.
To tackle this challenge, we introduce the U-shaped Vision Mamba (UVM-Net), an efficient single-image dehazing network.
Inspired by the State Space Sequence Models (SSMs), a new deep sequence model known for its power to handle long sequences, we design a Bi-SSM block that integrates the local feature extraction ability of the convolutional layer with the ability of the SSM to capture long-range dependencies.
Extensive experimental results demonstrate the effectiveness of our method. Our method provides a more highly efficient idea of long-range dependency modeling for image dehazing as well as other image restoration tasks.
The URL of the code is \url{https://github.com/zzr-idam}.

\end{abstract}
\section{Introduction}
Haze is a common atmospheric phenomenon that interferes with people's daily life as well as their judgment of goals, and heavy haze can even impact traffic safety. 
For computer vision, haze reduces the quality of images in most cases. It affects the model reliability in advanced vision tasks, further misleading machine systems, such as autonomous driving. 
Therefore, to perform efficient image dehazing through algorithms is a necessary work.

Single image dehazing aims to estimate the sharp image given a hazy input, which is a highly ill-posed problem. Conventional approaches are physically inspired and apply various sharp image priors~\cite{he2010single,fattal2014dehazing,zhu2015fast,berman2016non} to regularize the solution space. 
However, these methods depend on manual priors, and real-world haze is more complex than the preset priors.
Recently, deep learning approaches have shown promising performance~\cite{cai2016dehazenet,ren2016single,zhang2018densely,li2017aod,ren2018gated,liu2019griddehazenet,chen2019gated,dong2020physics,deng2020hardgan,dong2020multi,qin2020ffa,wu2021contrastive,wang2021eaa,shao2020domain}, especially image dehazing methods based on Transformer's architecture~\cite{song2023vision}.
%
%
%

Although these approaches achieve state-of-the-art results on a single image dehazing task, their models usually have a large size of parameters and are computationally expensive.
Note that self-attention mechanisms are quadratically proportional to the size of the input image, thus making them resource-intensive, especially for hazy images with high resolution that are usually available. Therefore, how to have efficient modeling of long-range dependencies of images remains an unresolved issue. Recently, State Space Sequence Models (SSMs)~\cite{GuThesis,SSM-NeurIPS21}, especially Structured State Space Sequence Models (S4) ~\cite{S4-ICLR22}, have emerged as efficient building blocks (or layers) for constructing deep networks, and have gained promising performances in the analysis of continuous long sequence data~\cite{SSM4Audio,S4-ICLR22}.
Recently, mamba~\cite{mamba} demonstrated excellent modeling capabilities for long-range dependencies, and it performed well for image~\cite{S4ND-neurips22} and video~\cite{SSM4Video-ECCV22} classification, and medical image segmentation~\cite{ma2024u,wang2024mamba}.

Inspired by U-Mamba~\cite{ma2024u} and Mamba-UNet~\cite{wang2024mamba}, we present a U-shaped Vision Mamba (UVM-Net), which forms a deep network based on the U-Net structure with both local capture capability and efficient long-range modeling.
Unlike U-Mamba and Mamba-UNet, we propose the Bi-SSM module, where we scroll the feature maps over the channel domain to fully utilize the long-range modeling capability of SSM.
U-Mamba and Mamba-UNet do not built a long-range dependency on another dimension of the feature map (the non-channel domain).
Extensive experimental results demonstrate that our method performs well on the task of image dehazing.
UVM-NET paves the way for future advances in network design that can show additional capabilities in other image restoration tasks.

\begin{figure}[t]
	\begin{center}\scriptsize
		\tabcolsep 1pt
		\begin{tabular}{@{}c@{}}
                \includegraphics[width = 0.95\textwidth]{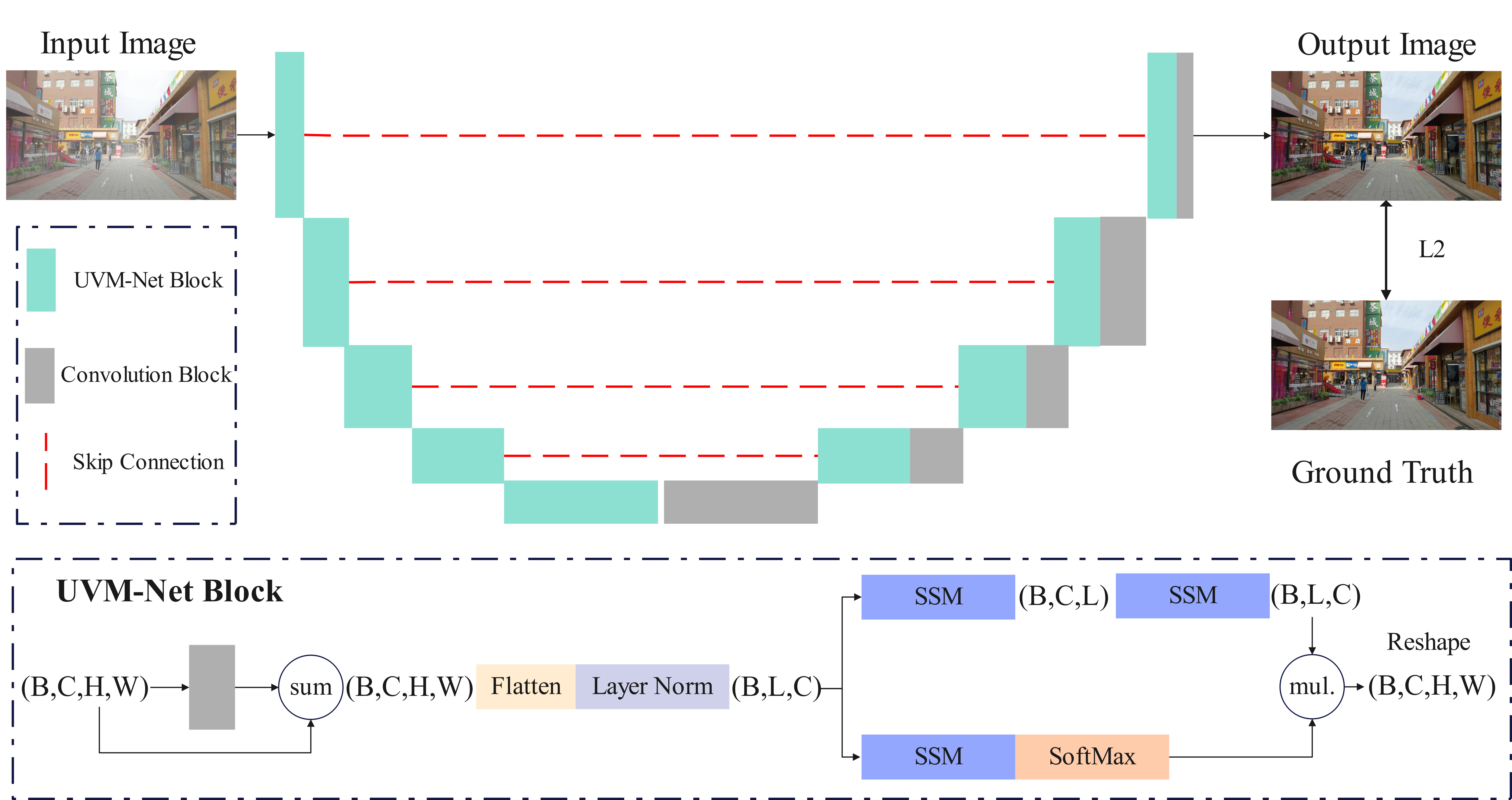}  
		\end{tabular}
	\end{center}
 \vspace{-2mm}
 	\caption{\textbf{Overview of the UVM-Net architecture.} UVM-Net employs the encoder-decoder framework with UVM-Net blocks in the encoder and convolution blocks in the decoder, together with skip connections. In UVM-Net block, our feature maps are first applied to a convolution operation, then the unfolded pixels are modeled over SSM, and the size of the final feature is reshaped to the size of the input information.}
	\label{f2}
 \vspace{-4mm}
\end{figure}

\section{Method}
UVM-Net follows the encoder-decoder network architecture that captures both local features and long-range information efficiently. Fig.~\ref{f2} shows an overview of the UVM-Net block and the whole network architecture. Next, we first introduce the Mamba block followed by illustrating the details of UVM-Net.

\begin{table*}[t] \tiny
  \centering
  \caption{
    Quantitative comparison of various dehazing methods trained on the RESIDE datasets.
  }
  \vspace{-4mm}
  \label{tab:quantitative}
  \begin{center}
    \renewcommand\arraystretch{1.25}
    {
      \begin{tabular}{|c|cc|cc|cc|cc|cc|}
        \hline
        \multirow{3}*{Methods}&\multicolumn{2}{c|}{ITS}&\multicolumn{2}{c|}{OTS}&\multicolumn{2}{c|}{RESIDE-6K} &\multicolumn{2}{c|}{RS-Haze} & \multicolumn{2}{c|}{\multirow{2}{*}{Overhead}}\\
        \cline{2-9}
         &\multicolumn{2}{c|}{SOTS-indoor}&\multicolumn{2}{c|}{SOTS-outdoor} &\multicolumn{2}{c|}{SOTS-mix} &\multicolumn{2}{c|}{RS-Haze-mix} & \multicolumn{2}{c|}{}\\
        \cline{2-11}
                                                            & PSNR  & SSIM  & PSNR  & SSIM   & PSNR  & SSIM  & PSNR  & SSIM  & \#Param & MACs \\
        \hline
        \hline
        (TPAMI'10) DCP~\cite{he2010single}                  & 16.62 & 0.818 & 19.13 & 0.815 & 17.88 & 0.816 & 17.86 & 0.734 & -       & -      \\
        (TIP'16) DehazeNet~\cite{cai2016dehazenet}          & 19.82 & 0.821 & 24.75 & 0.927 & 21.02 & 0.870 & 23.16 & 0.816 & 0.009M & 0.581G \\
        (ECCV'16) MSCNN~\cite{ren2016single}                & 19.84 & 0.833 & 22.06 & 0.908 & 20.31 & 0.863 & 22.80 & 0.823 & 0.008M & 0.525G \\
        (ICCV'17) AOD-Net~\cite{li2017aod}                  & 20.51 & 0.816 & 24.14 & 0.920 & 20.27 & 0.855 & 24.90 & 0.830 & 0.002M & 0.115G \\
        (CVPR'18) GFN~\cite{ren2018gated}                   & 22.30 & 0.880 & 21.55 & 0.844 & 23.52 & 0.905 & 29.24 & 0.910 & 0.499M & 14.94G \\
        (WACV'19) GCANet~\cite{chen2019gated}               & 30.23 & 0.980 & -     & -     & 25.09 & 0.923 & 34.41 & 0.949 & 0.702M & 18.41G \\
        (ICCV'19) GridDehazeNet~\cite{liu2019griddehazenet} & 32.16 & 0.984 & 30.86 & 0.982 & 25.86 & 0.944 & 36.40 & 0.960 & 0.956M & 21.49G \\
        (CVPR'20) MSBDN~\cite{dong2020multi}                & 33.67 & 0.985 & 33.48 & 0.982 & 28.56 & 0.966 & 38.57 & 0.965 & 31.35M & 41.54G \\
        (ECCV'20) PFDN~\cite{dong2020physics}               & 32.68 & 0.976 & -     & -     & 28.15 & 0.962 & 36.04 & 0.955 & 11.27M & 50.46G \\
        (AAAI'20) FFA-Net~\cite{qin2020ffa}                 & 36.39 & 0.989 & 33.57 & \textbf{0.984} & 29.96 & 0.973 & 39.39 & 0.969 & 4.456M & 287.8G \\
        (CVPR'21) AECR-Net~\cite{wu2021contrastive}         & 37.17 & 0.990 & -     & -     & 28.52 & 0.964 & 35.69 & 0.959 & 2.611M & 52.20G \\
        DehazeFormer-T~\cite{song2023vision}                            & 35.15 & 0.989 & 33.71 & 0.982 & 30.36 & 0.973 & 39.11 & 0.968 & 0.686M & 6.658G \\
        DehazeFormer-S~\cite{song2023vision}                           & 36.82 & 0.992 & 34.36 & 0.983 & 30.62 & 0.976 & 39.57 & 0.970 & 1.283M & 13.13G \\
        DehazeFormer-B~\cite{song2023vision}                           & 37.84 & 0.994 & \textbf{34.95} & \textbf{0.984} & 31.45 & 0.980 & 39.87 & 0.971 & 2.514M & 25.79G \\
        DehazeFormer-M~\cite{song2023vision}                           & 38.46 & 0.994 & 34.29 & 0.983 & 30.89 & 0.977 & 39.71 & 0.971 & 4.634M & 48.64G \\
        DehazeFormer-L~\cite{song2023vision}                           & 40.05 & \textbf{0.996} & -     & -     & -     & -     & -& -& 25.44M & 279.7G \\
        Ours                                     & \textbf{40.17} & \textbf{0.996} & 34.92     & \textbf{0.984}     & \textbf{31.92}     & \textbf{0.982}     & \textbf{39.88}& \textbf{0.972} & 19.25M & 173.55G \\
        \hline
      \end{tabular}
    }
  \end{center}
\end{table*}

\subsection{UMV-Net}
Mamba has demonstrated impressive results on a variety of discrete data, but it remains underexplored in the field of image dehazing. Here, we take advantage of Mamba's linear scaling to enhance the long-range dependency modeling of traditional U-Net.
%
%
%

As shown in Fig.~\ref{f2}, each building block contains two successive convolution blocks~\cite{resnet} followed by a UVM-net block. 
The convolution block contains the standard convolutional layer ($\times 3$) with Leaky ReLU~\cite{leakyRelu}. 
Image features with a shape of $(B, C, H, W)$ are then flattened and transposed to $(B, L, C)$ where $L=H\times W \times D$. After passing the Layer Normalization~\cite{layerNorm}, the features enter the Mamba block~\cite{mamba} that contains two parallel branches. In the first branch, the features are expanded to $(B, L, C)$ by a  SSM layer, next, the roll transforms into an $(B, C, L)$ before entering an SSM. 
In the second branch, the features are also expanded to $(B, L, C)$ by a  SSM layer, next, the feature map goes through softmax to form an attention map. Then, the features from the two branches are merged with the Hadamard product. Finally, the features are projected back to the original shape $(B, L, C)$ and then reshaped and transposed to $(B, C, H, W)$.
The entire model down-sampling and up-sampling of the feature map is consistent with standard U-Net, notably the bilinear interpolation used for up-sampling.


\section{Experiments}
The setup of the experiment follows exactly Dehazeformer's~\cite{song2023vision} configuration in the paper, including the number of iterations, the learning rate, and the batch size.
This is because it is currently the SOTA in the field of single-image dehazing and our structure is similar to it.
For this paper, we used PyTorch 1.8 and the GPU was a 3090RTX shader with 48G RAM.
We use the number of parameters (\#Param) and multiply-accumulate operations (MACs) to measure the overhead.
MACs are measured on $256\times256$ images.

\subsection{Quantitative Comparison}

We quantitatively compare the performance of UVM-Net and baselines, and the results are shown in TABLE~\ref{tab:quantitative}.
Here we bold the results where UVM-Net exceeds them.
Overall, our proposed UVM-Net outperformed these baselines.

\subsection{Qualitative Comparison}
We compared with dehazeformer on the publicly available dataset and the real-world image with haze, where dehazeformer is over-enhanced on the publicly available dataset.
\begin{figure*}[!t]
  \centering
  \includegraphics[width=1.0\textwidth]{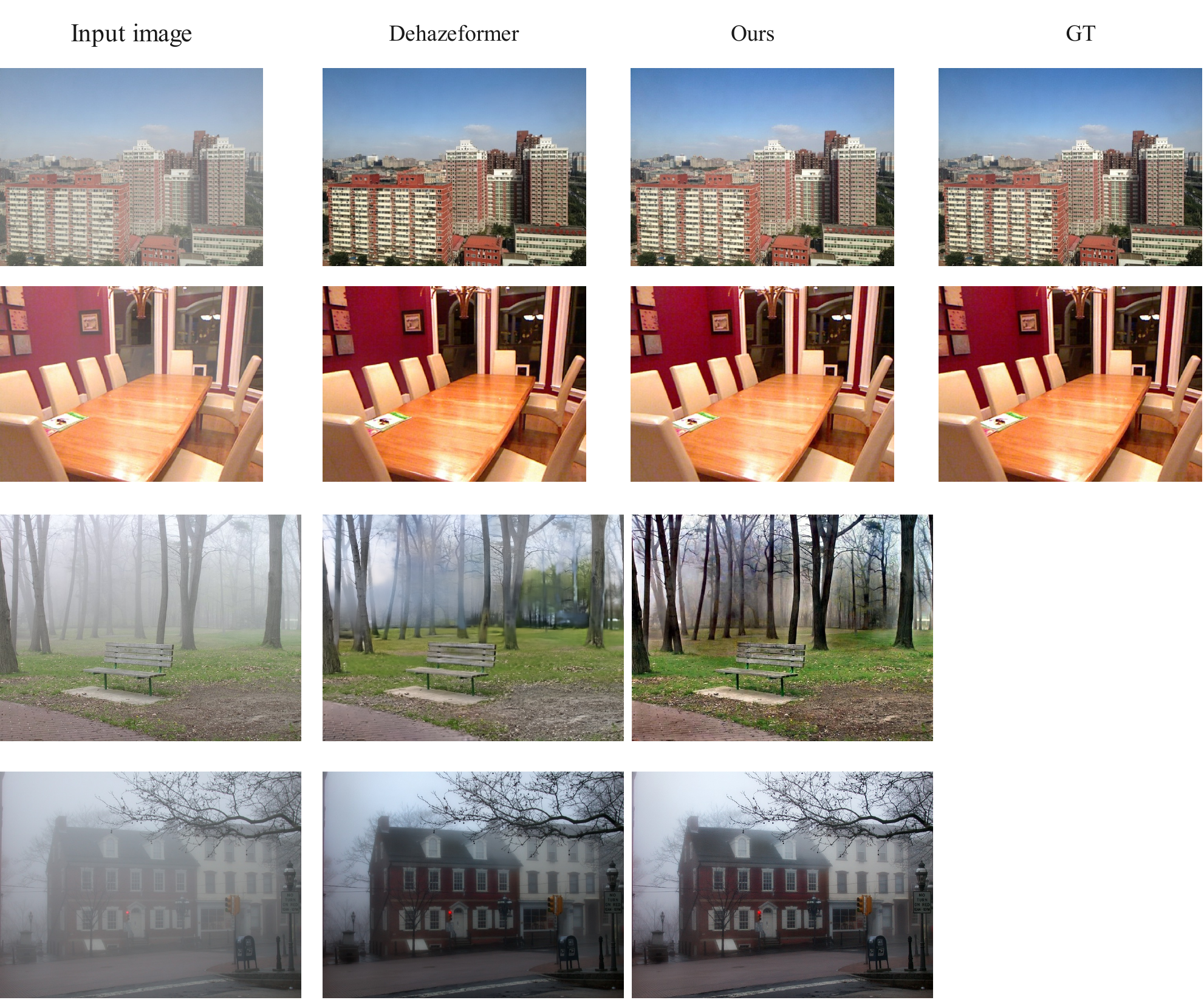}
  \caption{
    Qualitative comparison of image dehazing methods on SOTS mix set, where the first rows are outdoor images, and the second row is indoor images.
    The third and fourth rows are real-world images.
    The first column is the hazy images and the last is the corresponding ground truth.
  }
  \label{fig:compare1}
\end{figure*}

\subsection{Ablation Study}
We perform ablation studies on the RESIDE-Full's indoor scene.
We conducted two ablation experiments where 1) we removed the SSM to use 1D convolution instead, and 2) we removed the SSM to use scaled dot product instead.
The experimental results show the effectiveness of our method.
\begin{table}[h]
  \centering
  \caption{
    Ablation study on SSM module.
  }
  \label{tab:ablation4}
  \begin{center}
    \renewcommand\arraystretch{1.25}
      {
      \begin{tabular}{|l|cccc|}
        \hline
        Setting & PSNR & SSIM  & \#Param & MACs  \\
        \hline
        \hline
         Ours & 39.88 & 0.972 & 19.25M & 173.55G \\
        1D conv& 35.11 & 0.950 & 10.66M & 88.93G \\
        SDP & 38.25 & 0.968 & 23.88M & 229.93G \\
        \hline
      \end{tabular}
      }
  \end{center}
\end{table}

\begin{table*}[ht]
	\centering
	\caption{Compare quantization with the state-of-the-art image enhancement methods in LOL and MIT-Adobe FiveK (MIT5K).}
	\resizebox{\textwidth}{!}{
		\begin{tabular}{cc|cccccccccccc}
			\toprule
			\multicolumn{2}{c|}{} & BIMEF & CRM & LIME & RetinexNet & MBLLEN & DSLR & DRBN & ZeroDCE++ & KinD++ & TFR & Ours \\
			\midrule
			\multirow{3}{*}{LOL}   & PSNR \textcolor{red}{$\uparrow$} & 13.88 & 17.20 & 16.76 & 16.77 & 17.56 & 18.24 & 20.13 & 19.43 & 21.30  & 22.92  & \textbf{24.32}      \\
			& SSIM \textcolor{red}{$\uparrow$} & 0.577 & 0.644 & 0.564 & 0.567 & 0.736 & 0.787 & 0.802 & 0.768 & 0.822  & 0.837  & \textbf{0.859}      \\
			& NIQE \textcolor{blue}{$\downarrow$} & 7.69  & 8.02  & 9.13  & 9.73  & \textbf{3.46}  & 4.11  & 4.63  & 7.79 & 5.11 & 4.03 & 3.92      \\
			\midrule
			\multirow{3}{*}{MIT5K} & PSNR \textcolor{red}{$\uparrow$} & 18.67 & 13.99 & 11.21 & 20.81 & 16.42 & 17.02 & 20.95 & 16.46 & 22.01  & 25.03  & \textbf{25.78}      \\
			& SSIM \textcolor{red}{$\uparrow$} & 0.693 & 0.674 & 0.667 & 0.687 & 0.851 & 0.750 & 0.794 & 0.766 & 0.832  & 8.521  & \textbf{0.925}      \\
			& NIQE \textcolor{blue}{$\downarrow$} & 3.87  & 4.22  & 4.50  & 4.48  & 4.19 & 3.90  & 5.44  & 3.92 & 4.15 & 3.49 & \textbf{3.40}    \\
			\bottomrule 
	\end{tabular}}
	\vspace{-4mm}
	\label{tab:tab1}
\end{table*}

\begin{table*}[t!]
	\centering
	\caption{Comparison of quantitative results on five benchmark datasets. Best and second best values are indicated with \textbf{bold} text and \underline{underlined} text respectively.}
	\label{tab:result}
	\resizebox{\linewidth}{!}{
		\begin{tabular}{l|cccccccccc|cc}
			\bottomrule[0.5mm]
			Datasets  & \multicolumn{2}{c}{Test100}     & \multicolumn{2}{c}{Rain100H}     & \multicolumn{2}{c}{Rain100L}     & \multicolumn{2}{c}{Test1200}     & \multicolumn{2}{c|}{Test2800}    & \multicolumn{2}{c}{Average}     \\
			Methods   & PSNR           & SSIM           & PSNR           & SSIM           & PSNR           & SSIM           & PSNR           & SSIM           & PSNR           & SSIM           & PSNR           & SSIM           \\ \hline
			DerainNet~\cite{DerainNet} & 22.77          & 0.810          & 14.92          & 0.592          & 27.03          & 0.884          & 24.31          & 0.861          & 23.38          & 0.835          & 22.48          & 0.796          \\
			SEMI~\cite{semi}      & 22.35          & 0.788          & 16.56          & 0.486          & 25.03          & 0.842          & 24.43          & 0.782          & 26.05          & 0.822          & 22.88          & 0.744          \\
			DIDMDN~\cite{DIDMDN}    & 22.56          & 0.818          & 17.35          & 0.524          & 25.23          & 0.741          & 28.13          & 0.867          & 29.95          & 0.901          & 24.58          & 0.770          \\
			UMRL~\cite{UMRL}      & 24.41          & 0.829          & 26.01          & 0.832          & 29.18          & 0.923          & 29.97          & 0.905          & 30.55          & 0.910          & 28.02          & 0.880          \\
			RESCAN~\cite{RESCAN}    & 25.00          & 0.835          & 26.36          & 0.786          & 29.80          & 0.881          & 31.29          & 0.904          & 30.51          & 0.882          & 28.59          & 0.857          \\
			PReNet~\cite{PReNet}    & 24.81          & 0.851          & 26.77          & 0.858          & 32.44          & 0.950          & 31.75          & 0.916          & 31.36          & 0.911          & 29.42          & 0.897          \\
			MSPFN~\cite{MSPFN}     & 27.50          & 0.876          & 28.66          & 0.860          & 32.40          & 0.933          & 32.82          & 0.930          & 32.39          & 0.916          & 30.75          & 0.903          \\
			MPRNet~\cite{MPRNet}    & 30.27          & 0.897          & 30.41          & 0.890          & 36.40          & 0.965          & 33.64          & 0.938          & 32.91          & 0.916          & 32.73          & 0.921          \\
			KiT~\cite{KiT}       & 30.26          & 0.904          & 30.47          & 0.897          & 36.65          & 0.969          & 33.85          & 0.941          & 32.81          & 0.918          & 32.81          & 0.929          \\
			DGUNet~\cite{DGUNet}    & 30.32          & 0.899          & 30.66          & 0.891          & 37.42          & 0.969          & 33.68          & 0.938          & 33.23          & 0.920          & 33.06          & 0.923          \\
			IDT~\cite{IDT}       & 29.69          & 0.905          & 29.95          & 0.898          & 37.01          & 0.971          & 33.38          & 0.937          & 31.38          & 0.908          & 32.28          & 0.924          \\
			HINet~\cite{HINet}     & 30.26          & 0.905          & 30.63          & 0.893          & 37.20          & 0.969          & 33.87          & 0.940          & 33.01          & 0.918          & 33.00          & 0.925          \\
			DCT~\cite{DCT}       & 30.91          & 0.912          & \underline{30.74} & 0.892 & 38.19          & \underline{0.974} & \underline{33.89} & \underline{0.941} & \underline{33.57} & \underline{0.926} & 33.46          & \underline{0.929} \\
			SFNet~\cite{SFNet}     & \underline{31.47} & \underline{0.919} & \textbf{31.90} & \underline{0.908} & \underline{38.21} & \underline{0.974} & 33.69          & 0.937          & 32.55          & 0.911          & \underline{33.56} & \underline{0.929} \\ \hline
			Ours      & \textbf{32.11} & \textbf{0.915} & 30.66          & \textbf{0.909}          & \textbf{39.66} & \textbf{0.985} & \textbf{34.92} & \textbf{0.942} & \textbf{33.89} & \textbf{0.952} & \textbf{33.98} & \textbf{0.940} \\ \bottomrule[0.5mm]
	\end{tabular}}
\end{table*}

\section{Other task}
\noindent \textbf{Low light enhancement.} 
We evaluate the performance of our method on two paired public datasets, including the LOL dataset \cite{Chen2018Retinex} and the MIT-Adobe FiveK dataset \cite{bychkovsky2011learning} (see Table~\ref{tab:tab1}). 
We randomly select 450 pairs divided in the dataset for training and the other 50 pairs for testing. 
Each pair contains a low-light image and its corresponding well-exposed reference image. 
MIT-Adobe FiveK contains 5000 images recorded with a DSLR camera, and the tonal attributes of all images are manually adjusted by five photographic experts (A/B/C/D/E).
Reference with~\cite{W-Box}, we use the retouched results of expert C as the ground truth. 
We manually divide the MIT-Adobe FiveK dataset into two parts at a ratio of 8:2, using the first 4000 images to train the model and randomly selecting 500 images from the remaining images for testing.
We uniformly convert the raw images in the above dataset to PNG format for training and testing.
The Implementation of the environment uses the PyTorch~\cite{pytorch} open-source framework, trained and tested on Intel(R) Xeon(R) Gold 5218 CPU @ 2.30GHz, 128G RAM, and TITAN RTX3090 GPU with 24G RAM. 
We set the batch size to 12 and trained our model using the AdamW optimizer with $\beta_1$ = 0.9, $\beta_2$ = 0.999, and $\epsilon = 10^{-8}$. 
The learning rate is initialized to 0.0001, and 200 epochs are enforced, decaying by a factor of 0.5 in every 50 epochs.

\noindent \textbf{Deraining.} 
we conduct extensive experiments on the Rain13K training dataset which contains 13700 clean/rain image pairs. For the testing, five synthetic benchmarks (Test100~\cite{Test100}, Rain100H~\cite{Rain100HL}, Rain100L~\cite{Rain100HL}, Test2800~\cite{Test12002800}, and Test1200~\cite{Test12002800}) are considered for evaluation. For each image used, we use BLIP to generate corresponding text descriptions.
We calculate the PSNR and SSIM along the Y channel in the YCbCr color space as quantitative comparisons. We compare our method with 14 image deraining approaches, including DerainNet~\cite{DerainNet}, SEMI~\cite{semi}, DIDMDN~\cite{DIDMDN}, UMRL~\cite{UMRL}, RESCAN~\cite{RESCAN}, PReNet~\cite{PReNet}, MSPFN~\cite{MSPFN}, MPRNet~\cite{MPRNet}, KiT~\cite{KiT}, DGUNet~\cite{DGUNet}, IDT~\cite{IDT}, HINet~\cite{HINet}, DCT~\cite{DCT} and SFNet~\cite{SFNet}. In Tab.~\ref{tab:result}, we report the quantitative comparison results among different methods.
We conduct experiments in PyTorch on NVIDIA GeForce RTX 3090 GPUs. The network is trained for a total of 300k iterations with a batch size of 8. The initial learning rate is $2\times10^{-4}$, and the Adam optimizer is used to decay with the cosine annealing schedule. During training, we utilize cropped patches of size 128 $\times$ 128 as input, and to augment the training data, random horizontal and vertical flips are applied to the input images.

\section{Discussion}
Our model is a potentially large visual model (3.9G weight file).
Limited by the consumer GPU's RAM, the fixed weight of 64 can be increased to 256 if the number of GPUs is sufficient.

\section{Conclusion}
In this paper, we present a new architecture, UVM-Net, for single image dehazing, which integrates the advantages of local mode recognition of CNN and global context understanding of SSM. The results show that UVM-Net has the promise to be the backbone of promising image restoration networks for the next generation.

\bibliographystyle{splncs04}
\bibliography{ref}

\end{document}